# Kolmogorov-Arnold Networks for metal surface defect classification

**Maciej Krzywda[*], Mariusz Wermiński[2], Szymon Łukasik[*3,6] and Amir H. Gandomi[4,5]**

[*] *AGH University of Krakow,*
*Faculty of Physics and Applied Computer Science,*
*al. A. Mickiewicza 30, 30-059 Krakow, Poland*
*e-mail: krzywda@agh.edu.pl*
[2] *AGH University of Krakow*
*Faculty of Metals Engineering and Industrial Computer Science*
*al. A. Mickiewicza 30, 30-059 Krakow, Poland*
*e-mail: marwer@agh.edu.pl*
[3] *Systems Research Institute, Polish Academy of Sciences, ul. Newelska 6, 01-447 Warsaw, Poland*
[4] *Faculty of Engineering and IT, University of Technology Sydney, 5 Broadway, Ultimo NSW 2007, Australia*
[5] *University Research and Innovation Center (EKIK), Óbuda University, Bécsi út 96/B, Budapest, 1034, Hungary*
[6] *NASK National Research Institute, ul. Kolska 12, Warsaw 01-045, Poland*

**Abstarct:** *This paper presents the application of Kolmogorov-Arnold Networks (KAN) in classifying metal surface defects. Specifically, steel surfaces are analyzed to detect defects such as cracks, inclusions, patches, pitted surfaces, and scratches. Drawing on the Kolmogorov-Arnold theorem, KAN provides a novel approach compared to conventional multilayer perceptrons (MLPs), facilitating more efficient function approximation by utilizing spline functions. The results show that KAN networks can achieve better accuracy than convolutional neural networks (CNNs) with fewer parameters, resulting in faster convergence and improved performance in image classification.*

## Sieci Kolmogorov-Arnold w klasyfikacji defektów powierzchniowych metali

**Streszczenie:** *W niniejszej pracy przedstawiono zastosowanie sieci Kolmogorov-Arnold (KAN) w klasyfikacji defektów powierzchni metali. W szczególności badane są powierzchnie stali pod kątem wykrywania takich wad, jak pęknięcia, wtrącenia, łaty, powierzchnie z wżerami i zarysowania. Sieci KAN, oparte na twierdzeniu Kolmogorova-Arnolda, stanowią innowacyjną alternatywę dla tradycyjnych wielowarstwowych perceptronów (MLP), umożliwiając efektywniejsze aproksymowanie funkcji poprzez zastosowanie funkcji sklejanych. Wyniki badań wskazują, że sieci KAN mogą osiągać lepszą dokładność niż konwolucyjne sieci neuronowe (CNN) przy mniejszej liczbie parametrów, co skutkuje szybszą zbieżnością i lepszymi wynikami w klasyfikacji obrazów.*

## 1. Introduction

The field of deep learning is rapidly evolving, with continuous advancements in neural network architectures significantly contributing to progress in the image classification field [1,2,3]. Convolutional Neural Networks (CNNs) have become a cornerstone in analyzing multidimensional data, such as images, due to their capability to automatically extract meaningful features from raw data. In recent years, there has been a growing integration of advanced mathematical theories into deep learning architectures[4], enhancing neural networks' ability to process complex data structures. Among the promising alternatives to traditional Multilayer Perceptron (MLPs), Kolmogorov-Arnold Networks (KANs) leverage the Kolmogorov-Arnold theorem and utilize splinefunction as a key element of their architecture. Inlight of these developments, this workexplores the adaptation of KAN to convolutional layers, commonly used in CNN architectures for image classification. One such practical application of these networks is the classification of surface defectsof metallic parts, as steel and other alloys play a pivotal role in various industries, including automotive, defense, and machinery manufacturing. However, themanufacturing process of such materials often faces challenges related to quality control, leading to the occurrence of various defects, such as cracks, inclusions, patches, pitting surfaces, and scratches [5,6]. Effective classification of these defects is essential for ensuring the high quality of metallic partsproduction.The rationale behind utilizing KAN architectures, which incorporate learnable activation functions along edges, lies in their ability to increase both expressive capacity and efficiency. By substituting linear weight matrices with spline functions, KANs significantly reduce the number of required parameters to attain high accuracy, resulting in quicker convergence and improved generalization performance.

## 2. Kolmogorov-Arnold Networks

KANs are novel of neural network architecture based on the Kolmogorov-Arnold representation theorem [6,7]. They provide an alternative way to approximate functions by using learnable, parameterized univariate functions as activation functions rather than fixed ones typically used in MLPs. The Kolmogorov-Arnold theorem, a significant result in mathematical analysis, states that any continuous multivariate function can be expressed as a finite sum of continuous univariate functions. This gives KAN architectures a solid theoretical foundation, suggesting that complex multivariate functions can be broken down into simpler, univariate components.

## 3. Comparision of KAN and MLP

The advantages of spline functions (B-spline) over traditional activation functions in neural networks, particularly in the context of Kolmogorov–Arnold Networks (KAN), are quite distinctive.In particular, KANs offer significant differences in neural network construction that allow for greater flexibility and interpretability of models.
Major differences between MLP and B-spline based networks are:

- Better representation of local dependencies: Spline functions allow precise fitting to local data due to their structure, enabling accurate modeling of univariate functions, as described in the Kolmogorov-Arnold representation theories. Traditional activation functions like ReLU or Sigmoid lack this flexibility in local fitting, which can cause difficulties in accurately modeling complex data structures, especially with a small number of parameters [6,8]
- Reduction of the curse of dimensionality: KANs utilize spline functions that effectively address the challenge of high-dimensional data, making these models more scalable in solving regression tasks. In contrast, MLPs, particularly those using classic activation functions, can struggle with this issue[8,9]
- Adaptability of the B-spline grid: The spline functions used in KANs are parameterized as adaptable grids. This means that KANs are able to adjust themselves to changing input data during the learning process, whereas traditional activation functions in MLPs remain static and do not offer this level of flexibility[6]
- Better performance with limited data: Thanks to spline functions, KANs can provide better results with smaller datasets, as their structure is more efficient at uncovering internal data relationships. MLPs with traditional activation functions, such as ReLU, may require significantly larger datasets to achieve comparable performance [6,8].

In practice, KANs apply this theorem[6,7] by constructing neural networks that learn these univariate functions. The architecture usually consists of an input transformation layer, where each input variable passes through a learnable univariate function. This is followed by a summation layer that aggregates the outputs of these functions, creating intermediate values. Subsequently, the output layerprocesses the sum through additional learnable univariate functions, and the final outputs are produced by summing their results. Essentially, KAN simplifies the task of approximating a complex function by breaking it down into smaller, manageable parts that are adjusted and learned during training.

In contrast, MLPs use fixed functions like ReLU, sigmoid, or tanh, which stay the same throughout the entire training process[10,11]. This rigidity can sometimes limit the MLP's ability to model more complex relationships. In terms of approximation power, KANs are theoretically capable of representing any continuous function on a compact domain, just like MLPs, which are known to be universal approximators when given enough width and depth. However, KANs might achieve the same level of function approximation with fewer parameters, making them a more efficient optionin some cases. Training these networks also differs from MLPs. While MLPs rely on well-established methods like backpropagation and gradient descent, KAN training involves learning the parameters of univariate functions, which can require more specialized optimization techniques.

## 4. Experiment settings

In our experiment, we decided to use two datasets of surface metal defects:

- **Neu Metal Surface Defects Data** [12] contains 1800 images of metal surfaces (200x200 pixels) with six defect classes: Crack, Inclusion, Patch, Pitted Surface, Rolled-in Scale, and Scratches.
- **Severstal** [13]: Steel Defect Detection includes 12,568 images of steel surfaces (256x1600 pixels) with four defect classes: Rolled-in Scale, Patch, Craze, and Pitted Surface.

In thisexperiment, thebaseline model of CNNs was built, utilizing commonly applied Convolutional and Linear layers typically used in classification tasks. For clarity and ease of reference,the specific namesofthese models have been assigned, which are detailed in the following description:

- **TwoLayerConvNet**: Two convolutional layers with five filters, ReLU activations, followed by max-pooling and a final fully connected layer for classification.
- **TwoLayerConvNetPlus**: Two convolutional layers with five and twenty five filters, ReLU activations, max-pooling, and two fully connected layers for more complex pattern learning**.**
- **SingleLayerLinearNet**: A single fully connected layer applied to the flattened input for simple classification.
- **FourLayerConvNet**: Four convolutional layers with increasing filter sizes, ReLU activations, max-pooling, followed by two fully connected layers for classification.
- **TwoLayerConvKAN**: Two convolutional layers with ReLU activations, followed by a KAN Linear layer for enhanced feature transformation.
- **FourLayerConvKAN**: Four convolutional layers with ReLU activations, followed by a KANLinear layer for flexible input-output mapping.
- **ThreeLayerConvTwoLayerKAN**: Three convolutional layers with ReLU activations, followed by two KANLinear layers for flexible decision-making and regularization.

## 5. Results

Due to the limitation of using GPU for KAN networks, all models were running on CPU. EachKAN model has been trained for 100 epochs and all experiments was re-run 10 times.The comparisonof these results with baseline methods [14] ispresented in Table 2.

Based on the experimental results (Table 1, Figure 1 and Figure 2), itcan observed that the accuracy achieved by the KAN-based neural network is noticeably higherthan CNN. Moreover, this improvement is achieved with fewer neural network parameters.

***Table 1.*** *Classification Performance model comparison for the best network*

| Dataset | Model | Test Accuracy | Training Time (s) | Params |
|---|---|---|---|---|
| Severstal | TwoLayerConvNet | 0.76 | 73057 | 18374 |
| | ThreeLayerConvTwoLayerKAN | 0.79 | 120188 | 1054480 |
| | TwoLayerConvNetPlus | 0.78 | 81349 | 5020174 |
| | SingleLayerLinearNet | 0.74 | 77005 | 172804 |
| | FourLayerConvNet | 0.79 | 115963 | 14830372 |
| | TwoLayerConvKAN | 0.78 | 105067 | 270370 |
| | FourLayerConvKAN | 0.79 | 118189 | 248672 |
| NEU Metal Surface Defects | TwoLayerConvNet | 0.91 | 20452 | 75376 |
| | ThreeLayerConvTwoLayerKAN | 1.00 | 33741 | 2694800 |
| | TwoLayerConvNetPlus | 0.97 | 20970 | 14748688 |
| | SingleLayerLinearNet | 0.37 | 17457 | 720006 |
| | FourLayerConvNet | 0.97 | 32335 | 41045286 |
| | TwoLayerConvKAN | 0.93 | 28244 | 1125370 |
| | FourLayerConvKAN | 0.98 | 31529 | 889952 |

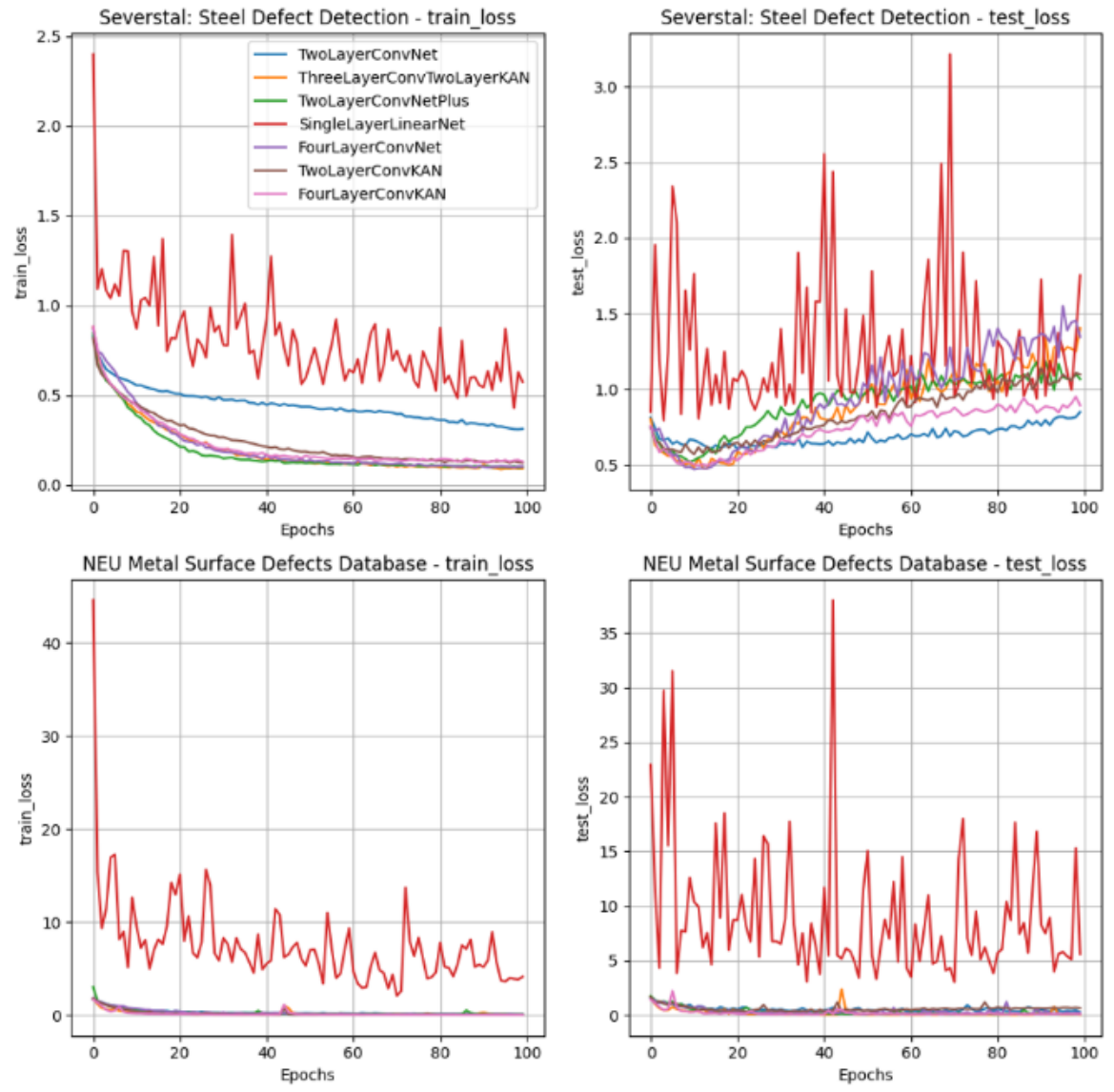

***Figure 1.*** *Loss Functionresults for presented best models*

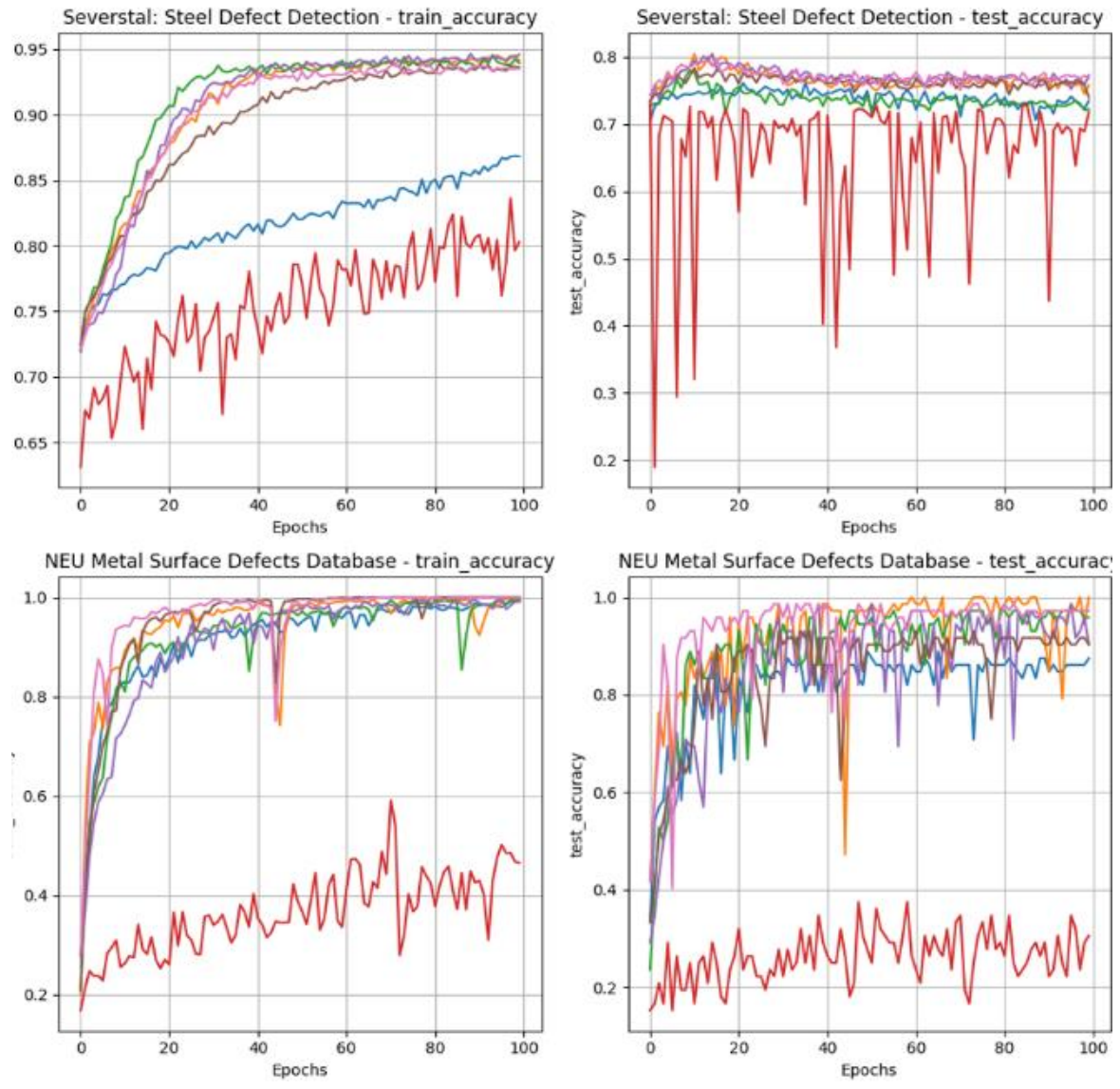

***Figure 2.*** *Test and Train Accuracyresults for presented best models*

Along with findings from similar research [15], the KAN-based network for image classification demonstrates promising outcomes. However, KANs are primarily designed for scenarios where high accuracy and interpretability are key objectives. While interpretability is indeed essential, for example, in large language models (LLMs) [16], it may have very different implications when compared to its meaning in scientific applications.

Regarding Table 1., Table 2. and the Severstal dataset, the **ThreeLayerConvTwoLayerKAN** model achieves one of the highest test accuracy scores at 0.79. However, this comes at the expense of a significantly longer training time and a large parameter count exceeding 1 million. The complexity of this model appears to aid in more accurate metal surface defect detection compared to simpler models, but it significantly increases computational time. The **TwoLayerConvKAN** model appears from the visual data to perform comparably to other KAN models, indicating that this architecture is effective in defect detection. Similarly, the **FourLayerConvKAN**model reaches a high test accuracy, nearing 0.98, but with fewer parameters than more complex models like **TwoLayerConvNetPlus**, showing a favorable balance between efficiency and performance. In the NEU Metal Surface Defects dataset, KAN-based models, compared to the results of the models in [14], perform exceptionally well, achieving near-perfect accuracy. The **ThreeLayerConvTwoLayerKAN** model achieves excellent accuracy, close to 1.0. However, its parameter count (2,694,800) and extended training time suggest it may be excessive for more straightforward tasks. The **FourLayerConvKAN** model maintains similarly high accuracy, close to 1.00, while using fewer parameters, making it a better balance between performance and model complexity. Overall, KAN architecture models deliver superior accuracy and defect classification effectiveness. On the Severstal dataset, they achieve some of the highest scores, though simpler models such as **TwoLayerConvNetPlus** (Table 1) also perform very well, suggesting that these KAN-based models are most suited for more complex tasks. However, the computational costs associated with training these models are considerable, which may be a limitation for resource-constrained applications.

***Table 2**. Juxtaposition of KAN modelsperformancewith baselines from [14]*

| Dataset | Model | Test Accuracy |
|---|---|---|
| Severstal | TwoLayerConvKAN | 0.775 +/- 0.019 |
| | FourLayerConvKAN | 0.788 +/- 0.006 |
| | ThreeLayerConvTwoLayerKAN | 0.793 +/- 0.005 |
| NEU Metal Surface Defects | TwoLayerConvKAN | 0.93 +/- 0.032 |
| | FourLayerConvKAN | 0.977 +/- 0.006 |
| | ThreeLayerConvTwoLayerKAN | 0.99 +/- 0.01 |
| | CNN [14] | 93.24 |
| | KNN [14] | 83.22 |
| | Siamese neural network [14] | 28.22 |

## 6. Conclusions

This research shows great promise and yields satisfactory results; however, a significant challenge lies in the lengthy training times. In our study, we observed how the KAN-based network performed when applied to a smaller dataset. Even with relatively small KAN-based neural networks, the computational demands were substantial, and with larger datasets and more complex networks, the training time could become prohibitively long. This limitation may lead to abandoning this approach in favor of architectures specifically designed for such problems.

## Acknowledgement

This work was partially supported by the program 'Excellence Initiative - Research University' for the AGH University of Krakow and by a Grant for Statutory Activity from the Faculty of Physics and Applied Computer Science of the AGH University of Krakow.We gratefully acknowledge Polish high-performance computing infrastructure PLGrid (HPC Center: ACK Cyfronet AGH) for providing computer facilities and support within computational grant no. PLG/2023/016643.

.